# Attention-LSTM for Multivariate Traffic State Prediction on Rural Roads

Elahe Sherafat[1], Bilal Farooq[1], Amir Hossein Karbasi[2] and Seyedehsan Seyedabrishami[3]

[1]Laboratory of Innovations in Transportation, Toronto Metropolitan University, Toronto, ON, Canada.
[2]Department of Civil Engineering, McMaster University, Hamilton, ON, Canada.
[3]Faculty of Civil and Environmental Engineering, Tarbiat Modares University, Tehran, Iran.

Contributing authors: esherafat@torontomu.ca; bilal.farooq@torontomu.ca; karbaa3@mcmaster.ca; seyedabrishami@modares.ac.ir;

**Abstract**

Accurate traffic volume and speed prediction have a wide range of applications in transportation. It can result in useful and timely information for both travellers and transportation decision-makers. In this study, an Attention based Long Sort-Term Memory model (A-LSTM) is proposed to simultaneously predict traffic volume and speed in a critical rural road segmentation which connects Tehran to Chalus, the most tourist destination city in Iran. Moreover, this study compares the results of the A-LSTM model with the Long Short-Term Memory (LSTM) model. Both models show acceptable performance in predicting speed and flow. However, the A-LSTM model outperforms the LSTM in 5 and 15-minute intervals. In contrast, there is no meaningful difference between the two models for the 30-minute time interval. By comparing the performance of the models based on different time horizons, the 15-minute horizon model outperforms the others by reaching the lowest Mean Square Error (MSE) loss of 0.0032, followed by the 30 and 5-minutes horizons with 0.004 and 0.0051, respectively. In addition, this study compares the results of the models based on two transformations of temporal categorical input variables, one-hot or cyclic, for the 15-minute time interval. The results demonstrate that both LSTM and A-LSTM with cyclic feature encoding outperform those with one-hot feature encoding.

**Keywords:** Traffic state prediction, A-LSTM, deep learning, rural roads

# 1   Introduction

Several major problems are associated with transportation networks, such as traffic congestion, road safety, and high travel time variability. The development of Intelligent Transportation Systems (ITS) was aimed at solving these problems, operating as a vital part of traffic management and control [1]. As a result of advancements in technology and the development of big data, Artificial Intelligence (AI) methods, particularly deep learning methods, have become an essential part of ITS for traffic management. A practical application of AI is the prediction of short-term traffic patterns and speeds, which is the basis for modern traffic management [2, 3].

Short-term traffic flow and speed prediction allows traffic departments to gain accurate and timely information to intervene ahead of time. Besides, travellers are able to set their departure time and plan their trip with increased reliability, which leads to alleviating congestion. Moreover, to improve traffic safety, short-term traffic flow prediction is of great social and economic importance [4]. Therefore, traffic flow and speed prediction models improve safety and reduce congestion and its negative consequences, such as air pollution [5].

Due to the importance of traffic flow and speed prediction, many researchers have presented several models, such as statistical, machine learning, and deep learning models, to predict traffic characteristics with high accuracy [1, 3, 6–10]. Some of these studies predict traffic flow [1, 11, 12] and others speed [13, 14]. There are a few studies that predict traffic flow and speed simultaneously [2, 15]. Besides, there can be seen several gaps in terms of simultaneous prediction of traffic flow and speed.

These studies have mostly predicted traffic flow and speed in freeways and urban networks [2, 12, 13] and only a few of them aim at rural roads [16, 17]. However, accurate prediction of traffic characteristics in some rural segmentations is important and challenging due to the following reasons. Firstly, the travel behaviour on these roads differs from that of urban roads and depends more on calendar variables. For example, congestion on rural roads deteriorates during the holidays [18]. Hence, the rural networks require specific traffic analysis and prediction of their own.

The second reason is the traffic safety problem on these roads since they account for many accident fatalities. For instance, rural roads accounted for 54% of all fatalities in the U.S. in 2012, despite 19% of the population living in rural areas [19]. Thus, there is a need to develop models for accurate prediction of traffic flow and speed, specifically on rural roads.

To the best of the authors' knowledge, no study has predicted both flow and speed on rural roads using a multivariate deep-learning model. Therefore, this study aims to simultaneously predict traffic flow and speed on a rural highway to fill these gaps. In this study, multivariate deep neural networks, established on Attention-Based Long Short-Term Memory (A-LSTM) and Long Short-Term Memory (LSTM) artificial neural networks, are presented to predict traffic flow and speed simultaneously. The time-series cross-validation technique is deployed for training and validation of the models using the 5, 15, and 30 minutes time intervals. Besides, the data analysis is conducted to investigate traffic behaviour in the rural segmentation of the case study.

As a case study, we utilized the data from the Karaj-Chalus rural highway, a two-lane, two-way road connecting Tehran to Chalus in Iran. This rural highway has recurrent issues, which make it essential to control the traffic. Chalus is one of Iran's most popular holiday destinations due to its pleasant weather and natural attractions. This rural highway can become highly congested and even experiences blockage during the holidays as people from Tehran flock to Chalus [20]. This, in turn, deteriorates the air pollution problem and damages the environment.

Another problem is road safety, as several sharp curves go through the rain forests and the narrowness and steep mountainous terrain, making the road potentially dangerous [21]. In 2016, there were 855 road crashes with 37 deaths on Karaj-Chalus rural highway [21]. Hence, accurate traffic flow and speed prediction are essential to address these problems proactively.

The main contributions of this study are as follows:

- Developing a multi-output A-LSTM to simultaneously predict traffic flow and speed on a rural highway.
- Comparing the performance of A-LSTM and LSTM in predicting traffic flow and speed.
- Investigating the optimal temporal aggregation level for the model and its effect on predictive performance.
- Based on the optimal time interval, comparing the performance of two neural networks with the one-hot and cyclic transformation of time-series categorical input variables.

The rest of this paper is organized as follows. Section 2 reviews the previous studies that deployed different methods to predict traffic flow and speed. Section 3 explains the methodology; firstly, the preliminary notations are introduced in section 3.1. The structure of the proposed A-LSTM is introduced in Section 3.2, and the time-series cross-validation method is discussed in Section 3.3. In Section 4, the characteristics of the road segmentation, the dataset, and the extracted features deployed in the study are discussed. Section 5 demonstrates numerical experiments and results. Finally, Section 6 presents the conclusions and suggestions for future studies.

## 2  Literature Review

Previous studies have predicted traffic flow and speed using a variety of approaches such as statistical [6–8], machine learning [9, 10, 22–26], and deep learning approaches [2, 12, 14–17, 27–32]. Regarding statistical methods, Ahmed and Cook [6] first proposed an Auto-Regressive Integrated Moving Average (ARIMA) model to predict freeway short-term traffic flow. Moreover, there were other variants of the ARIMA model which aimed to improve the performance of traffic flow prediction. For example, considering the fact that the ARIMA model does not handle nonlinear traffic data, Van Der Voort et al. [7] presented the KARIMA model that combined the Kohonen network and ARIMA to solve the shortcoming of the ARIMA model by improving short-term traffic flow prediction performance. Williams and Hoel [8] predicted short-term traffic flow based on a seasonal ARIMA (SARIMA) model.

Statistical approaches utilize simple models with high computational complexity, making them suitable for smooth and small sample data [33, 34]. However, since traffic data contain nonlinear and stochastic properties, statistical models perform unacceptably and consequently make traffic prediction unreliable as a result [35]. In contrast, conventional machine learning methods are able to capture better complex and nonlinear patterns in traffic data than statistical models [36].

Concerning machine learning approaches, Support Vector Machine (SVM) and Support Vector Regression (SVR) models have been the most widely deployed in previous studies. Their results showed that SVM and SVR models outperformed statistical models [10, 22]. For example, Hong et al. [10] combined a genetic algorithm and SVR model to create a model predicting traffic flow which outperformed the SARIMA model.

Moreover, Hu et al. [22] deployed a hybrid Particle Swarm Optimization (PSO)-SVR model for predicting traffic flow. This model is capable of reducing model learning time by processing noisy data effectively. They showed that the accuracy of the PSO-SVR model is higher than other models, such as SVM and ARIMA. with regard to traffic speed prediction, Wang and Shi [9] developed a hybrid model called C-WSVM for forecasting short-term traffic speed based on SVM, regression theory, and chaos-wavelet Analysis. Their results illustrated that although there is no significant difference between the performance of the C-WSVM and SVM models, the C-WSVM model's performance is better when the traffic states change. Generally, machine learning models showed better performance than statistical models.

In recent years, several studies have predicted traffic flow and speed based on deep learning methods. On the one hand, the volume of traffic data has increased due to the development of data collection technologies [3]. On the other hand, deep learning methods have the potential for analyzing and fitting big data with complexity and nonlinearity [1]. This encouraged many researchers to apply deep learning methods for short-term traffic flow and speed prediction.

Due to the time-series nature of traffic flow and speed data, Recurrent Neural Networks (RNNs) have been widely used to predict traffic flow and speed. These networks benefit from temporal memory functions, which make them useful for dealing with time-series data [3]. Although RNNs struggle with gradient vanishing problems, there are variants of them that can solve this problem. Take LSTM and Gated Recurrent Units (GRU) neural networks as examples. [37] utilized LSTM neural networks to predict traffic flow in freeway networks. Their results show that the performance of the LSTM model is better than the SVR model.

Moreover, some studies improved LSTM models. For instance, Ma et al. [12] presented a model with an upgraded LSTM to improve short-term urban road traffic flow prediction accuracy. First, this model has done a time-series analysis on traffic flow data in order to create a reliable time-series [38]. Then traffic flow data were fed into the upgraded model based on LSTM and bidirectional LSTM neural networks. The results of this study illustrated that the proposed model outperformed other models, such as the LSTM.

Tran et al. [14] proposed a deep learning approach that utilizes an LSTM network with a hyper-parameter tuning in urban arterial roads to predict short-term traffic speeds in parallel multi-lane roads. They [14] showed that the performance of the updated version of LSTM outperformed ARIMA, multi-layer perceptron (MLP), and Convolutional Neural Networks (CNNs) models.

Chen et al. [39] applied an attention-based LSTM model to predict traffic flow in freeway and highway networks. They showed that this model could improve the accuracy of traffic flow prediction. Wu et al. [13] proposed an attention-based LSTM model in order to predict traffic speed in urban networks. They demonstrated that the attention-based LSTM model could improve the accuracy of traffic speed prediction compared with the LSTM model.

In addition, based on multi-task learning models, Zhang et al. [2] presented a multitask learning model with GRU neural networks to predict traffic flow and speed simultaneously in a freeway network. Moreover, a multitask learning method based on graph convolutional networks (GCNs) and GRU neural networks were applied by Buroni et al. [15] to predict traffic flow and speed simultaneously on the freeway and urban road networks. The results of these studies illustrate that multi-task learning models have the potential to improve the accuracy of traffic flow and speed prediction.

Although several studies applied different methods to predict traffic flow and speed, some key gaps exist in previous studies. First, there are only a handful of studies that employed deep learning models to predict traffic flow and speed based on a multi-task learning framework. The further point is that these studies have used either freeway or urban traffic data, and to the best of our knowledge, no study has simultaneously predicted flow and speed based on rural road data.

In addition, it is important to apply traffic flow and speed prediction models on rural roads because they help solve safety and congestion issues that occur on rural roads, especially during holidays [18]. To fill these gaps, this study has employed neural network models based on LSTM and A-LSTM for multivariate prediction of traffic flow parameters, volume and average speed, on a rural highway.

In this study, extracted 5-minute time interval data and 15 and 30-minute aggregated data have been deployed for feature extraction, data analysis, training, and model evaluation. In addition, this study compared the performance of simple LSTM and A-LSTM based on the three horizons and time-series cross-validation. Furthermore, based on the best time interval, the performance of the neural network based on one-hot and cyclic feature encoding of the categorical variables is compared.

## 3   Methodology

Deep learning models are the most popular machine learning methods and have attracted researchers in industry and academia in the last decade. The reason is their potential for dealing with complex data with nonlinearity. RNNs are extensively deployed in traffic flow prediction due to their potential to capture the time-series nature of traffic data.

In this study, we have developed our proposed multivariate A-LSTM to simultaneously predict traffic volume and speed in a rural road segmentation. Besides, we trained a simple LSTM as a baseline to investigate the promise of the proposed A-LSTM deep neural network.

In this section, we first introduce the notation and define the problem in Section 3.1. Then the proposed A-LSTM structure and its mathematics are investigated in Section 3.2. The time-series cross-validation method used for model training and evaluation is explained in Section 3.3.

The general framework of the study is shown in Figure 1. The raw data are preprocessed, and handcrafted explanatory features are added to the dataset. In this stage, the temporal variables are defined using either one-hot or cyclic encoding mechanisms. The 5-minute interval data are aggregated into 15 and 30-minute interval datasets. All three sequences are fed into both LSTM and A-LSTM to predict the volume and speed of the next time step. Besides, we deployed a time-series cross-validation method to split the sequences and model training and evaluation.

### 3.1   Preliminaries and notation

$Y_t$ (*V*, *S*) denotes the traffic state at the time step *t*, which is made up of the equivalent hourly volume (vehicle/hour) of vehicles passing a specific section of the road, *V*; and the average speed (kilometre/hour) of those vehicles, *S*, for the time interval *t*. For *i* historical time intervals, the traffic state historical data *Hist* can be denoted as $Hist = \{Y_{(t-i+1)}, Y_{(t-i+2)}..., Y_t\}$.

Our proposed network takes the volume and speed of five previous time steps, $H_{i_t} = \{Y_{t-5}, Y_{t-4}..., Y_{t-1}\}$ according to Figure 3. Vector $X_t \in R_N$ represents the feature vector for time *t* and includes features such as hour, season, and being a holiday, as explained in Table 2. The dimension $R$ differs based on the transformation method of temporal variables.

The network aims to predict the traffic features $V$ and $S$ while taking the historical flow data of previous time steps, $H_{i_t}$, and the input feature of the current time step, $X_t$, as the input.

## 3.2 Attention-Based Long Short Term Memory (A-LSTM) Structure

Generally, RNNs use building blocks at each time step that store hidden states, the memory of the network. Using the memory, RNNs can keep track of traffic information and temporal dependencies over time and predict the future traffic state. Although, the simple RNN has the shortcoming of capturing long-term dependencies due to the vanishing gradient problem. The LSTM variant proposed to address the vanishing gradient problem of the simple RNN [1]. LSTM neural networks are widely used and have shown great promise to predict traffic flow and speed due to their potential to take the time-series nature of the traffic parameters into account [40].

We deployed the LSTM module in the structure of our proposed model. The LSTM has the potential to take the historical data and capture the time dependencies. It can memorize and store the information in the cell state that connects hidden units in the sequence. The cell state is updated in each building block using the forget, input, and output gates. During the learning process, the weight and bias parameters of these gates are updated to generate the traffic state in the next time step, as shown in Figure 2.

These gates decide what new information should be added to the cell state, what needs to be deleted, and what the model will generate as the output. In other words, it has the ability to read, write, save and delete information on the cell state to best predict the traffic volume and speed in each time step based on the time series variables from previous ones and the input features from the current one.

To introduce our proposed model, we first explain the mathematics behind each gate in the LSTM building blocks 2. Then, we discuss how the attention mechanism improves the performance of the vanilla LSTM model. When the cell state enters a building block unit, the forget gate decides whether to save or delete information from previous time steps; that is to say, it selects the optimal time lag for the input sequences [41]. The forget gate takes current input $x_t$, concatenation of $H_{i_t}$ and $X_t$, and passes it through Equation 2 using the sigmoid function Equation 1 as is illustrated in Figure 2. The output updates the cell state using Equation 5.

$$\sigma(t) = \frac{1}{1+e^t} \qquad (1)$$

$$f_t = \sigma(W_f[h_{t-1}, x_t] + b_f), \qquad (2)$$

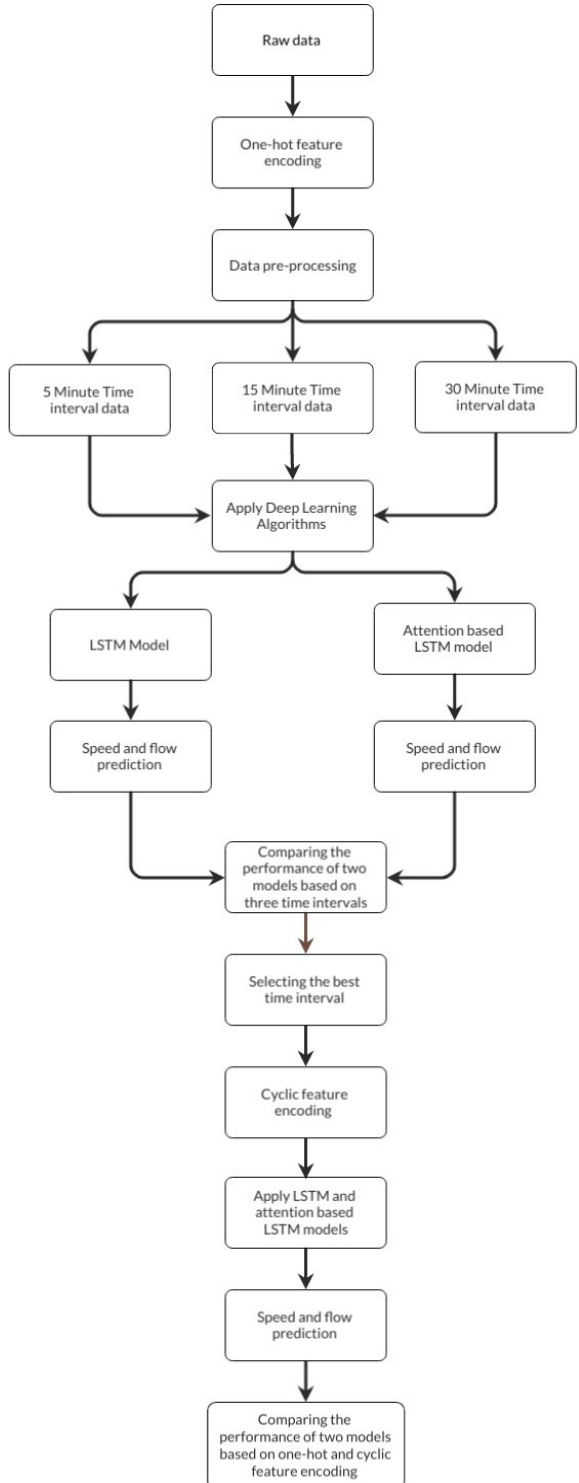

**Fig. 1**: The study framework

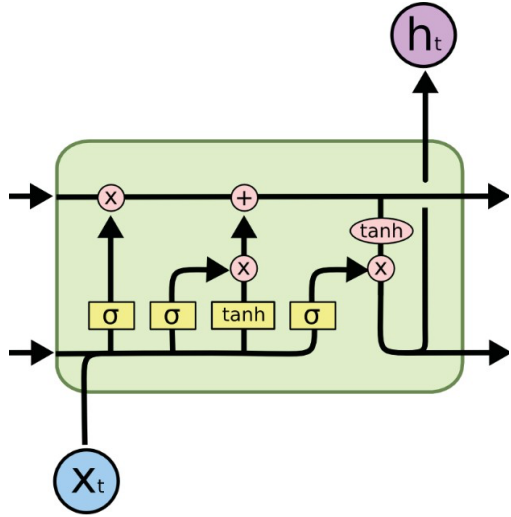

**Fig. 2**: A building block for a single time step in the LSTM architecture, comprising input, forget, and output gates.

Then the writing gate uses sigmoid and hyperbolic tangent as activation functions and specific weights and biases ($W_i$, $B_i$ & $W_c$, $b_c$) to store new information in the cell state based on Equations 3, 4, 5, and Figure 2.

$$i_t = \sigma\,(W_i\,[h_{t-1}, x_t] + b_i), \tag{3}$$

$$\tilde{C}_t = \tanh\,(W_c\,[h_{t-1}, x_t] + b_c), \tag{4}$$

$$C_t = f_t \times C_{t-1} + i_t \times \tilde{C}_t \tag{5}$$

The same concatenated input is passed through the sigmoid function, Equation 1 and parameters ($W_o$ and $b_o$) to create the hidden state of the current timestep $t$ based on Equations 6 and 7, Figure 2.

$$o_t = \sigma\,(w_o\,[h_{t-1}, x_t] + b_o), \tag{6}$$

$$h_t = o_t \times \tanh\,(C_t) \tag{7}$$

The hidden state provided in this section will be the argument for computing the alignment score in Equation 8 and the context vector in Equation 10 in the attention mechanism. The attention mechanism introduced by Bahdanau et al. [42] aimed at the fixed-length context vector problem in the RNNs. This restricts RNNs' predictive ability when dealing with long time-series sequences. The attention mechanism could be integrated into any RNN to improve its performance, especially when dealing with long sequences.

Simple RNNs and LSTM inherently have an encoding and decoding process to map the input sequence to the output. The encoder maps the input sequence to the context vector, and the decoder takes the context vector to produce the traffic flow parameters at time step *t*. When the attention mechanism is added to the architecture, the encoder assigns weights to each time step so that the generated context vector is a more efficient input for the decoder to generate outputs.

These weights are tuned in the learning process so that the network pays more attention (assigns more weight) to the information in a longer sequence, which is important to predict current volume and speed. E.g., we could see a specific pattern like existing a holiday just after a weekend in past sequences, and by deploying the attention layer, we can recall the pattern when it happens in the current time step. The computation process comprises three main computing steps: Alignment Scores, Weights, and Context vectors, Figure 3.

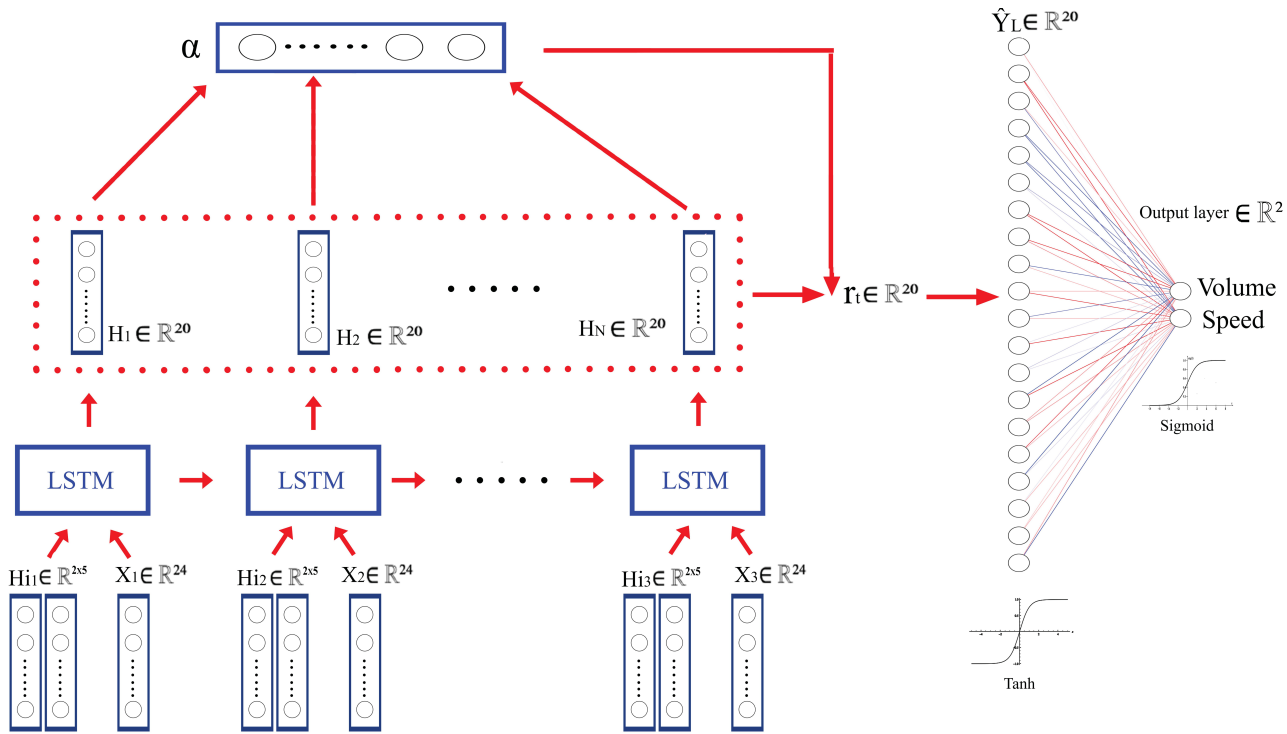

**Fig. 3:** The structure of proposed A-LSTM deep neural network

### 3.2.1 Alignment Scores

The alignment score, $e_{t,i}$, indicates how well the output $V_t$ and $S_t$ align with the input sequence elements. It is computed using the function $a$ () of time step $t$, which takes the decoder output of the previous time step, $o_{t-1}$, and the encoder hidden state, $h_i$, as arguments in a feed-forward process.

$$e_{t,i} = a\,(o_{t-1},\, h_i) \tag{8}$$

### 3.2.2 Weights

Weights, $\alpha_{t,i}$, are obtained from passing alignment scores computed from the previous step through a softmax function, Equation 9. The weights indicate where exactly the model should focus on the hidden state to better estimate the traffic state at time t. Figure 3 shows the $\alpha$ on the top of the hidden states, $H_i$.

$$\alpha = softmax\,(e_{t,i}) \tag{9}$$

### 3.2.3 Context Vector

Finally, the attention layer improves the model's performance by using the context vectors, $r_t$, instead of hidden states, $H_t$, to generate the output by the decoder, Figure 3. The context vector is computed as the weighted sum of all the hidden states over t, Equation 10.

$$r_t = \sum_{i=1}^{L} \alpha_{t,i} h_i \tag{10}$$

This context vector is deployed to compute the final output of the LSTM layer using the attention mechanism, $\hat{Y}_L = \{\hat{V}_t, \hat{S}_t\}$, using Equation 11.

$$\hat{Y} = f(V \times r_t + b_v) \tag{11}$$

The LSTM output would be represented as $\hat{Y}_L \in R^{20}$ and passes through the hyperbolic tangent function, Figure 3. Finally, two neurons of a dense layer will compute the normalized volume and speed using the sigmoid activation function. The attention mechanism incorporates into the LSTM architecture by mapping inputs to outputs in a forward direction during training, as discussed in detail. We deployed the 'Adam' optimizer and the 'mean square error' loss function to compile the model. For training the model, we chose a batch size of 128 based on a trial and error process. The total number of trainable parameters of the network is 4,463 parameters. We found 100 as the number of epochs that best guarantee learning and avoid overfitting based on results in Section 5 and Figures 8a to 8f.

We trained and evaluated our proposed A-LSTM model, Figure 3.2 and the vanilla LSTM model as a baseline to investigate the performance of the proposed model. The split method, input variables type, and the sequences' intervals are discussed in Section 3.3.

For the input selection, we relied on data analysis results as well as the trial and error procedure during training. Two representations of time-series variables were deployed as input variables. In the first scenario, the cyclic transformation of time-series features is deployed. Whereas the second uses the one-hot representation. These features, along with five previous time steps of output features, shaped the total number of 34 and 45 input variables in the cyclic and one-hot transformation scenarios respectively. Hence the number of input nodes and network structures differs based on the input dimension.

### 3.3 Time-Series Cross-Validation

Unbiased and robust validation is essential for evaluating the performance of a model. The cross-validation technique has demonstrated potential for tuning the hyper-parameters, estimating the performance of the models, and selecting the most generalized one [43]. K-fold cross-validation assumes observations are independent of each other and that there is no correlation between them through time.

On the other hand, the time-series cross-validation method takes the temporal dependencies of records into account. Hence, we employed the time-series cross-validation technique in this study to evaluate the performance of the models in predicting the average speed and volume while taking into account their time-series nature. Figure 4 illustrates three time-series cross-validation sets that separate training sets from validation and testing.

In this method, the origin of the validation-test sets rolls forward and moves the fixed length of it toward the end of the sequence. During this procedure, the size of train sets increases, but the validation-test set size remains the same. For example, for the 15-minute interval dataset, as demonstrated in Table 1, the train sizes are 16,175, 32,348, and 48,521 for the split sets, while both validation and test sets are of the same size of 8,086 in all three splits. That is to say, the percentage of the train set size increases from 64% in the first split to 75% in the last one for all intervals.

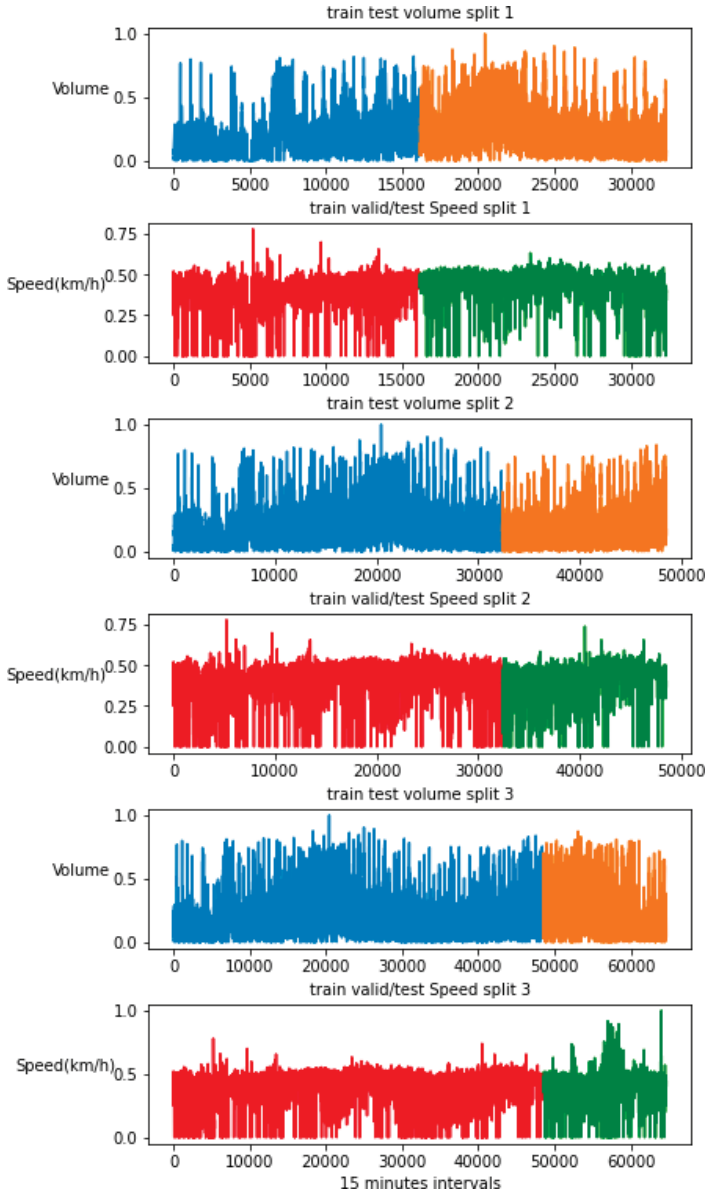

**Fig. 4**: Time-series cross-validation for splitting the train and validation-test dataset. volume train sets (blue), speed train sets (red), volume validation-test sets (orange), and speed validation-test sets (green), all for 15 minutes interval

Table 1: The sample size for time-series cross validation

| Horizon (minutes) | 5 | | | 15 | | | 30 | | |
|---|---|---|---|---|---|---|---|---|---|
| Set split | train | valid | test | train | valid | test | train | valid | test |
| 1 | 48091 | 24044 | 24044 | 16175 | 8086 | 8087 | 8118 | 4057 | 4057 |
| 2 | 96179 | 24044 | 24044 | 32348 | 8086 | 8087 | 16233 | 4057 | 4057 |
| 3 | 144267 | 24044 | 24044 | 48521 | 8086 | 8086 | 24348 | 4057 | 4057 |

# 4  Case Study

This section first describes the rural road segment and its characteristics in Section 4.1. Then, the dataset and extracted features used as input variables are introduced in Section 4.2.

## 4.1  Data and road segment introduction

The data deployed for training and evaluating the proposed models in this study are the 5-minute volume (vehicle/hour) and average speed (kilometre/hour) of all kinds of vehicles passing a critical[1] segmentation on a rural road located in the north of Iran. Log-lasting blockage traffic state occurs in this segmentation during holidays and weekends. Besides, the characteristics of the road segment lead to low resilience and ability to be recovered from the blockage state.

Since 2010, the loop detectors in the rural network of the country have been collecting traffic data and reporting them online. Although the clean and prepossessed data are openly accessed for one-hour time intervals on the Road Maintenance and Transportation Organization (RMTO) website [44], The clean and preprocessed 5-minute time interval data are not available.

Since in this study, we are interested in investigating the variation of traffic flow parameters within one hour, the 5-minute interval raw traffic data were obtained initially. The data include two years of traffic data, from the beginning of 2018 to the end of 2019, just before starting the influence of the Covid pandemic and imposed travel restrictions. After cleaning the raw data, the handcrafted features were added to the database, and aggregated data were prepared for 15 and 30 minutes intervals.

The specific segmentation is located on Chalus road and connects Chalus to Tehran. Tehran is a polluted and crowded metropolis with 11,800 residents per square kilometre. While, Chalus is the primary vacation destination city in the country that is located only 145 kilometres from Tehran, Figure 5. Hence, during the holidays and weekends, a considerable population flocks to this two-way road that doesn't have any physical barriers. This results in a long-lasting blockage condition that increases the travel time up to four times compared with free-flow travel time.

This situation makes policymakers block one flow direction to dedicate both lanes to the reverse one by doubling the capacity to alleviate the blockage state. This policy, in turn, leads to safety hazards. These uncertainties make accurate traffic flow prediction of this rural segmentation the most challenging among 3,000 road segmentations in the rural road network.

## 4.2  Handcrafted feature extraction

This study utilizes less than one-hour intervals, and explanatory features were added to the raw database. Input features include time-related (month, day, season, and hour), calendar (day of week, holiday), weather, and flow-related features. The features are introduced in Table 2.

---

[1]A critical road segmentation refers to long-lasting traffic congestion conditions existence in this segmentation.

Table 2: Explanatory features introduction

| Input Feature | Explanation |
|---|---|
| DateTime | Jalali and Lunar DateTime is added to extract time-related features, e.g. holidays from the available Gregorian DateTime in the raw dataset. |
| Month | 1-12; is used as both a **one-hot** and **cyclic** variable. |
| Day | 1-31; is used as both a **one-hot** and **cyclic** variable. |
| Season | Spring-summer-fall-winter; is used as both a **one-hot** and **cyclic** variable. |
| Day-of-week | Monday to Sunday; is used as both a **one-hot** and **cyclic** variable. |
| Hour-288 | 1-288; specify which 5-minute interval in a day; the record is used only as a **cyclic** variable. |
| Hour | 1-24; is used as both a **one-hot** and **cyclic** variable. |
| 7-21 | 1 if the interval is between 7 AM and 9 PM, 0 otherwise. |
| Day-night | 1 if the interval is placed in the daytime, 0 otherwise; based on sunset and sunrise. |
| Holiday | Calendar holiday-weekend holiday-not a holiday; based on both Jalali and Lunar calendar, since public holidays vary based on both calendars; a **one-hot** variable. |
| Weather | Rainy-sunny-snowy; a **one hot** variable. |
| Volume-reveres | The normalized volume (vehicle/hour) in the reverse direction. |
| Average-speed-reverse | The normalized average speed (kilometre/hour) in the reverse direction. |
| Double-capacity | 1 if the vehicles can ride in both lanes, and the opposite direction is not allowed to use the road, 0 otherwise. |
| One-way | 1 if the vehicles are not allowed to use the road for dedicating both lanes to the other direction. |

Traffic flow features of the opposite direction are extracted from the data fusion of detectors. Volume and speed for the Chalus-Tehran direction are set as target variables (due to less percentage of missing values), and those features for the opposite direction (Tehran-Chalus) are used as input variables. Table 3 shows the correlation between target variables and the volume and speed of the opposite direction. Besides, since to alleviate the blockage in some cases, the road facilitates the traffic only in one direction, the two variables *one-way* and *double capacity* are added to the dataset, as explained in Table 2. The box plots of Figure 6 illustrate how the target variable, volume, differs based on categories of each explanatory variable. The analysis is extensively discussed in Section 5.1.

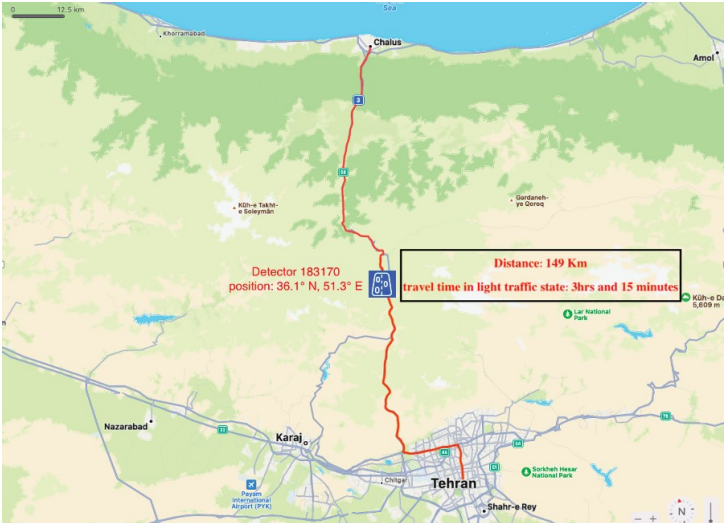

**Fig. 5**: The network demonstration and position of the detector. The detector is located in the path from Chalus to Tehran, just after the Kandovan tunnel. The volume and average speed data of the reverse direction (from Tehran to Chalus) have been used as input variables for the model.

## 5  Results

This section first illustrates the data analysis results and interprets them in Section 5.1. Then the results of the A-LSTM and LSTM models during training and evaluation for both input scenarios and different time horizons are illustrated and interpreted in Section 5.2.

### 5.1  Data Analysis Result

In this study, data analysis is conducted to select the significant input variables to be fed into the neural network model. Besides, deployed data analysis techniques address the black-box nature of neural networks by providing interpretable results. To this end, Correlation analysis between numeric inputs and target variables, volume and average speed is conducted as the results are illustrated in Table 3. Besides, based on Figure 6, box plots are deployed to investigate the relationship between categorical input variables and the continuous output volume.

For each categorical variable, the distribution of volume is analyzed using sets of side-by-side box plots for every level of each categorical variable. The box plots in Figure 6 demonstrate the distribution of volume (vehicle/hour) and take variables *Jalali day, month, hour, 7-21, day-night, season, weather,*

**Table 3**: Correlations between volume and speed for the target direction (Chalus-Tehran) and the reverse direction (Tehran-Chalus)

|  | Volume | Speed | Volume reverse | Speed reverse |
|---|---|---|---|---|
| Volume | 1 | 0.12 | 0.29 | -0.1 |
| Speed | 0.12 | 1 | 0.11 | 0.34 |
| Volume reverse | 0.29 | 0.11 | 1 | 0.15 |
| Speed reverse | -0.1 | 0.34 | 0.15 | 1 |

*holiday*, and *day-of-week* as a horizontal variable. The length and position of boxes of different categories for each categorical variable explain the difference in target variable distribution for each level of explenatory variables and the amount of explanation they provide.

Based on Figure 6, considering *Jalali day* variable, there can not be noticed much of differences in days of a Jalali month, so the input variable *Jalali day* is insignificant. *Jalali month* plot, on the other hand, demonstrates a higher amount of traffic volume in the first month and a continuous increase till the peak in months 4-6. This increase in the volume in these two periods is rooted in the new year holiday, which is 13 consecutive holidays, and the summer holiday, respectively. The *hour* box plot shows the peak hour is at 4 P.M., and the smooth curve demonstrates the meaningful and gradual changes in traffic during the day. The *7-21* and *day-night* variables' box plots also show tangible differences that these two variables can make; plots show a higher average and variation of traffic volume in a day than at night.

Regarding the *day-of-week* box plot, Friday and Saturday are experiencing the highest amount of traffic. This result is consistent with expectations since Thursdays and Fridays are the weekends in Iran. So it is expected to see a higher amount of traffic volume on Fridays and Saturdays when travellers are returning from their vacation to Tehran and Wednesdays and Thursdays in the opposite direction for travellers flocking toward their recreational destination (Chalus).

The *holiday*'s plot shows that weekends experience a higher average volume than calendar holidays and is found to be an informative variable. Although, for weather data, interpretation is challenging since it is expected that considering the danger that snow causes on any road with sharp curves and the hazard of avalanches, the average volume for the snow level category is less than others. In contrast, here, the reverse came true based on the plots. Although the analysis results are found incompatible with the common expectation, the variable *weather* is considered significant during the modelling process. Finally, variables in Table 2 are deployed in the final model training.

## 5.2 Neural Network Results for Predicting Traffic Flow Characteristics

This section demonstrates the results of multivariate traffic volume and average speed prediction using the LSTM and A-LSTM deep neural networks. Moreover, the impact of different time horizons is investigated. Finally, there

20     *Attention-LSTM for Multivariate Traffic State Prediction on Rural Roads*Table 4: Train, valid, and test set loss for LSTM and A-LSTM after 100 epochs

| | | MSE loss × 1000 | | | | | | | | |
|---|---|---|---|---|---|---|---|---|---|---|
| horizon(minutes) | | 5 | | | 15 | | | 30 | | |
| Set<br>Model | Split | Train | Valid | Test | Train | Valid | Test | Train | Valid | Test |
| LSTM | 1 | 41 | 65 | 40 | 25 | 43 | 26 | 33 | 61 | 31 |
| LSTM | 2 | 44 | 52 | 52 | 27 | 30 | 35 | 32 | 36 | 43 |
| LSTM | 3 | 44 | 65 | 60 | 27 | 44 | 36 | 32 | 50 | 47 |
| Average | all | 43 | 61 | 51 | 26 | 39 | 32 | 32 | 49 | 40 |
| Attention | 1 | 40 | 80 | 41 | 25 | 45 | 23 | 23 | 61 | 31 |
| Attention | 2 | 44 | 51 | 51 | 27 | 33 | 34 | 32 | 36 | 43 |
| Attention | 3 | 44 | 68 | 53 | 27 | 47 | 35 | 32 | 50 | 47 |
| Average | all | 43 | 66 | 48 | 26 | 42 | 30 | 29 | 49 | 40 |

will be a comparison between the performance of models based on cyclic and categorical time-series variables.

As mentioned earlier in Section 3.3, we conducted the training for three different split sets for unbiased validation. We trained A-LSTM and LSTm models over all the horizons- 5, 15, and 30 minutes. The mean square error loss function after 100 epochs is reported in Table 4.

The results show that A-LSTM outperforms LSTM concerning the testing sets loss function for the longer sequences, 5 and 15 minutes intervals. In contrast, for the validation sets of these two intervals, LSTM's performance is better. Regarding the 30-minute horizon, there is no meaningful difference between the two architectures. However, the difference between the models' performance increases as the time horizon decreases toward the 5-minute interval data. The reason is that the attention mechanism promises to perform better and mitigate the weakness of recurrent neural networks for dealing with longer sequences.

By comparing the performance of the models considering the time intervals of the input sequences, there can be seen that the 15-minute accounts for the minimum MSE with the value of 0.003 and 0.0032 for the A-LSTM and LSTM models, respectively. Figure 7 shows the 5-minute interval experiences more noise than other horizons. On the other hand, the 30-minute interval diagram experiences the minimum noise but loses more information within each time step compared with other horizons. Hence the 15-minute interval shows the best performance since neither has the high amount of noise as the 5-minute horizon has nor is too wide to lose a portion of the information as the 30-minute horizon is. The prediction and actual value diagrams follow the same trend through time in all three horizons for the A-LSTM model, Figure 7. This indicates the overall acceptable performance of the model.

Figure 8 demonstrates the MSE loss function decrease for 100 epochs during training for three different training and validation sets splits. According to the diagrams, each split shows a specific and unique behaviour regarding the decrease in loss values of the validation and training sets during the training process. That is to say, the splits coming from the time-series cross-validation method bring a significant difference in the evaluation of the model. Hence, the method results in an unbiased evaluation and has increased the model's generalization.

All the results above were related to the models that have taken cyclic time-series variables as inputs. Therefore, to investigate the impact of cyclic variables, this study also compares the results of the models based on both one-hot and cyclic time-series variables.

Table 5 compares results of models based on either one-hot or cyclic transformation of the temporal input variables. According to the average row, both LSTM and A-LSTM models perform better using cyclic variables for the testing set. However, A-LSTM experiences more improvement when using cyclic variables than the LSTM model. The average MSE loss function for the test set decreases from 33 to 32, while the same amount reaches 30 for the A-LSTM model using cyclic variables instead of one-hot ones.

**Table 5**: Train, valid, and test set loss for LSTM and A-LSTM after 100 epochs for the cyclic and one-hot transformation of input variables for 15-minute horizon

| | | MSE loss × 1000 | | | | | |
|---|---|---|---|---|---|---|---|
| Input data | | One-hot input variables | | | Cyclic input variables | | |
| Set Model | Split | Train | Valid | Test | Train | Valid | Test |
| LSTM | 1 | 25 | 42 | 31 | 25 | 43 | 26 |
| LSTM | 2 | 27 | 31 | 32 | 27 | 30 | 35 |
| LSTM | 3 | 27 | 45 | 36 | 27 | 44 | 36 |
| Average | 3 | 26 | 39 | 33 | 26 | 39 | 32 |
| Attention | 1 | 25 | 44 | 30 | 25 | 45 | 23 |
| Attention | 2 | 27 | 30 | 33 | 27 | 33 | 34 |
| Attention | 3 | 27 | 47 | 37 | 27 | 47 | 35 |
| Average | 3 | 26 | 40 | 33 | 26 | 41 | 30 |

Figure 9 demonstrates scatter plots that compare observed and calculated traffic parameters volume and average speed. Both models' diagrams are based on the third split of the A-LSTM. However, models in Figures 9a and 9b have taken cyclic variables as inputs. In contrast, those in Figures 9c and 9d are based on one-hot categorical variables. According to the figures, although both scatter plots are aligned with the 45-degree line, cyclic variable-based models fit and perform better than one-hot variables-based ones.

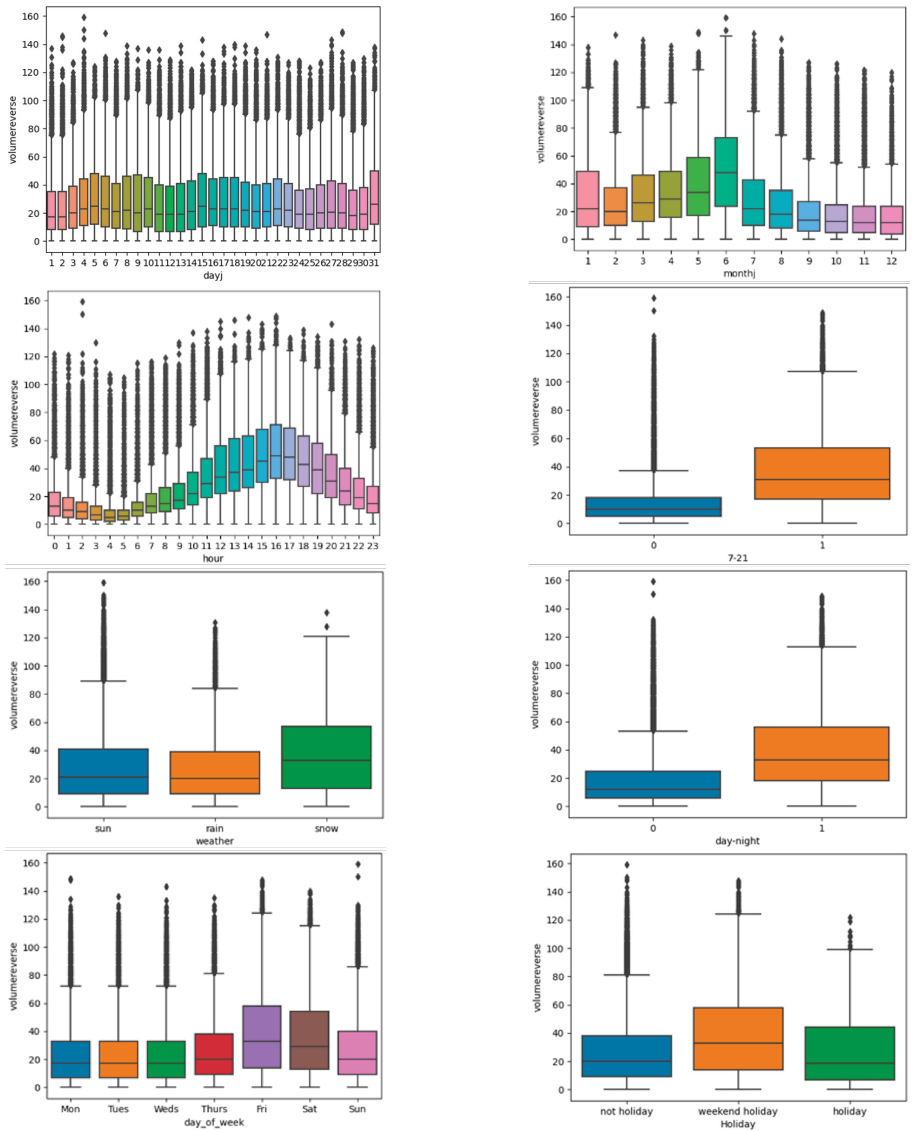

**Fig. 6**: The box plots of predictors' distribution with respect to volume predicted variable

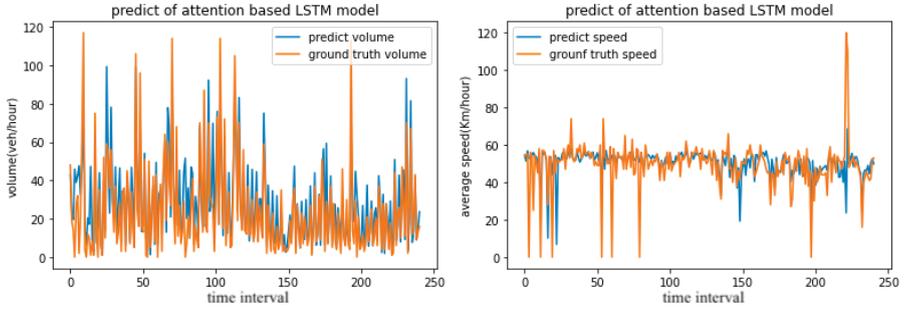

(a) Volume prediction for 5-minute time interval

(b) Average speed prediction for 5-minute time interval

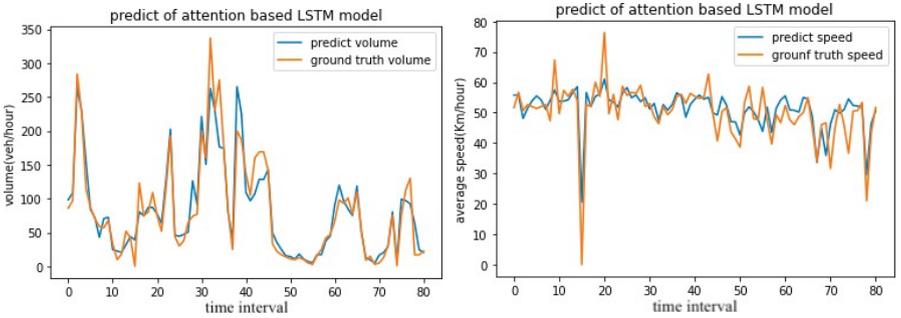

(c) Volume prediction for 15-minute time interval

(d) Average speed prediction for 15-minute time interval

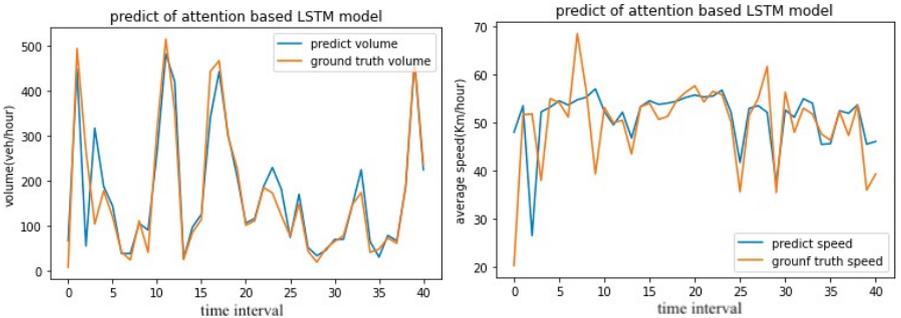

(e) Volume prediction for 30-minute time interval

(f) Average speed prediction for 30-minute time interval

**Fig. 7**: The plots of volume and average speed actual observed values (orange) versus the predicted values (blue) by the A-LSTM model for the third time-series cross-validation split on the testing dataset of the 5, 15, and 30-minute traffic flow dataset.

24        *Attention-LSTM for Multivariate Traffic State Prediction on Rural Roads*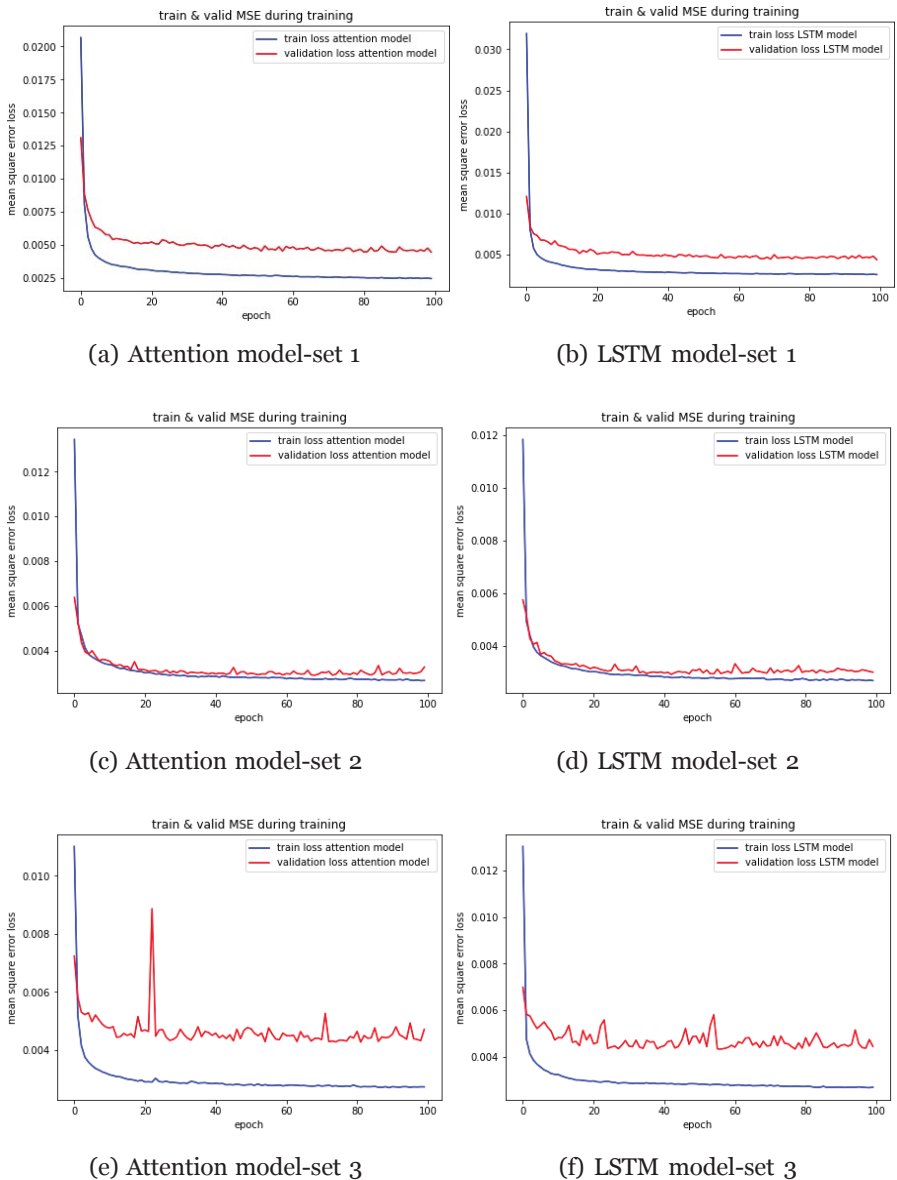

**Fig. 8**: Mean square error loss decrease during learning the LSTM (right side) and A-LSTM (left side) for time-series cross-validation sets

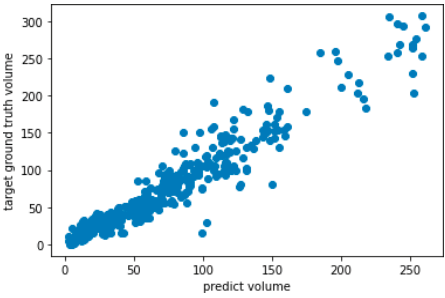
(a) Volume scatter plot for A-LSTM using cyclic variables

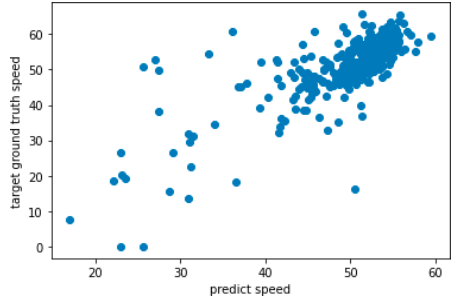
(b) Average speed scatter plot for A-LSTM using cyclic variables

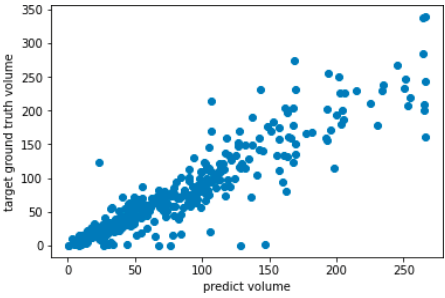
(c) Volume scatter plot for A-LSTM using one-hot categorical variables

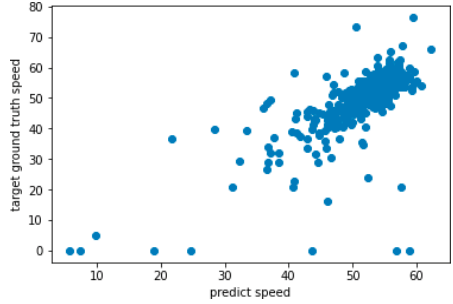
(d) Average speed scatter plot for A-LSTM using one-hot categorical variables

**Fig. 9**: The predicted versus true value for Volume (left side) and Average Speed (right side) for the third split of the A-LSTM model based on cyclic variables (on the top), and one-hot categorical variables (in the bottom)

# 6   Conclusion

Prediction of traffic flow parameters plays a vital role in Transportation Engineering and allows decision-makers to control and manage traffic in different transportation networks. On the other hand, traffic flow and speed prediction is an efficient tool that helps to reduce many traffic problems such as congestion and consequently other related problems such as air pollution and traffic safety issues. This paper aims to simultaneously predict traffic flow and speed on a rural road in Iran. The prediction of traffic flow and speed in this rural segmentation is crucial due to its unique characteristics, such as mountainous and dangerous routes and long-lasting congestion situations. This study aggregated 5-minute interval traffic data into 15 and 30-minute ones and deployed them for multivariate prediction of traffic flow and average speed using the LSTM and AB-LSMT models.

Generally, the results show the satisfying performance of both LSTM and A-LSTM for multivariate prediction of time-series traffic parameters, speed and volume, with a high level of nonlinearity and time dependencies. Moreover, the results show that based on the test dataset, the performance of A-LSTM is better than LSTM in 5 and 15-minute horizons. However, in the 30-minute time interval, there is no magnitude difference between the two deep learning models. It is worth mentioning that in the 5-minute time interval, the difference between the two models is increased, and the rationale reason is that the attention mechanism has the potential to improve performance and alleviate recurrent neural networks' weaknesses for dealing with longer sequences. Based on time interval comparison, the 15-minute horizon-based model performs best regarding two deep learning models' results.

It can be realized that the 15-minute time interval performs better than the 5 and 30-minute horizons because it has less noise than the 5-minute interval and, simultaneously, contains more information than the 30-minute interval. Finally, the study compares the performance of models based on two transformations of time-series input variables, cyclic and one-hot encoding. According to the results, models with cyclic variables outperform models with one-hot variables encoding.

In addition, for future studies, multivariate traffic parameter prediction of parallel paths using data from detectors in those segments and graph-based deep neural networks is recommended. Besides, the Spatiotemporal dependencies can be investigated using multiple detectors in one road. Future studies can work on the performance of LSTM and A-LSTM in other transportation networks, such as freeways and urban roads. Moreover, the different time interval comparison can be investigated in other networks, such as highways and freeways.

*Attention-LSTM for Multivariate Traffic State Prediction on Rural Roads*     27## Data Availability Statement

The traffic data from around 3,000 detectors on rural roads in Iran can be found open access on the Road Maintenance and Transportation Organization (RMTO) website [44] for one-hour time intervals. The open-access data include volume and average speed for each detector from 2010, as well as some extracted features. RMTO has not open-accessed the 5-minute data that were used in this study. The 5-minute raw data in 2018 came from detectors 183120 and 183170, which are located in opposite directions at the same station.